\documentclass[letterpaper, 10 pt, conference]{ieeeconf}
\IEEEoverridecommandlockouts
\usepackage[pdftex]{graphicx}

\usepackage[font=footnotesize]{subcaption}
\usepackage[font=footnotesize]{caption}

\usepackage[nobreak]{cite}
\usepackage[super]{nth}
\usepackage[ruled,linesnumbered]{algorithm2e}
\SetAlCapFnt{\footnotesize}
\SetAlgorithmName{Alg.}{Alg.}{List of Algorithms}

\usepackage{cancel} % cancel slash
\usepackage{times} % times symbol
\usepackage{amssymb, amsmath} % additional math functions
\usepackage{url}

\let\labelindent\relax %<-this prevents a conflict with IEEE \labelindent when using enumitem
\usepackage[inline]{enumitem}\setlist[enumerate]*{label=(\roman*)}

\usepackage{multirow}

\newcommand\figref[1]{\text{Fig.~\ref{#1}}}
\newcommand\tableref[1]{\text{Table~\ref{#1}}}
\newcommand\equref[1]{\text{Eq.~(\ref{#1})}}

\newcommand*{\sref}[1]{\S\ref{#1}}

\title{\LARGE \bf
Deep Segmented DMP Networks for Learning Discontinuous Motions
}
\author{Edgar Anarossi$^{1}$, Hirotaka Tahara$^{1}$, Naoto Komeno$^{1}$ and Takamitsu Matsubara$^{1}$% <-this % stops a space
\thanks{$^{1}$Authors are with the Division of Information Science, Graduate School of Science and Technology, Nara Institute of Science and Technology, Japan.}%
\thanks{This work was supported by JSPS KAKENHI Grant Numbers JP21H04910.}%
}

\begin{document}
\maketitle
\thispagestyle{empty}
\pagestyle{empty}

\begin{abstract}
Discontinuous motion which is a motion composed of multiple continuous motions with sudden change in direction or velocity in between, can be seen in state-aware robotic tasks.
Such robotic tasks are often coordinated with sensor information such as image.
In recent years, Dynamic Movement Primitives (DMP) which is a method for generating motor behaviors suitable for robotics has garnered several deep learning based improvements to allow associations between sensor information and DMP parameters.
While the implementation of deep learning framework does improve upon DMP's inability to directly associate to an input, we found that it has difficulty learning DMP parameters for complex motion which requires large number of basis functions to reconstruct. 
In this paper we propose a novel deep learning network architecture called Deep Segmented DMP Network (DSDNet) which generates variable-length segmented motion by utilizing the combination of multiple DMP parameters predicting network architecture, double-stage decoder network, and number of segments predictor.
The proposed method is evaluated on both artificial data (object cutting \& pick-and-place) and real data (object cutting) where our proposed method could achieve high generalization capability, task-achievement, and data-efficiency compared to previous method on generating discontinuous long-horizon motions.
\end{abstract}

\section{INTRODUCTION}

\begin{figure}[t]
\centering
    \includegraphics[width=1.0\hsize]{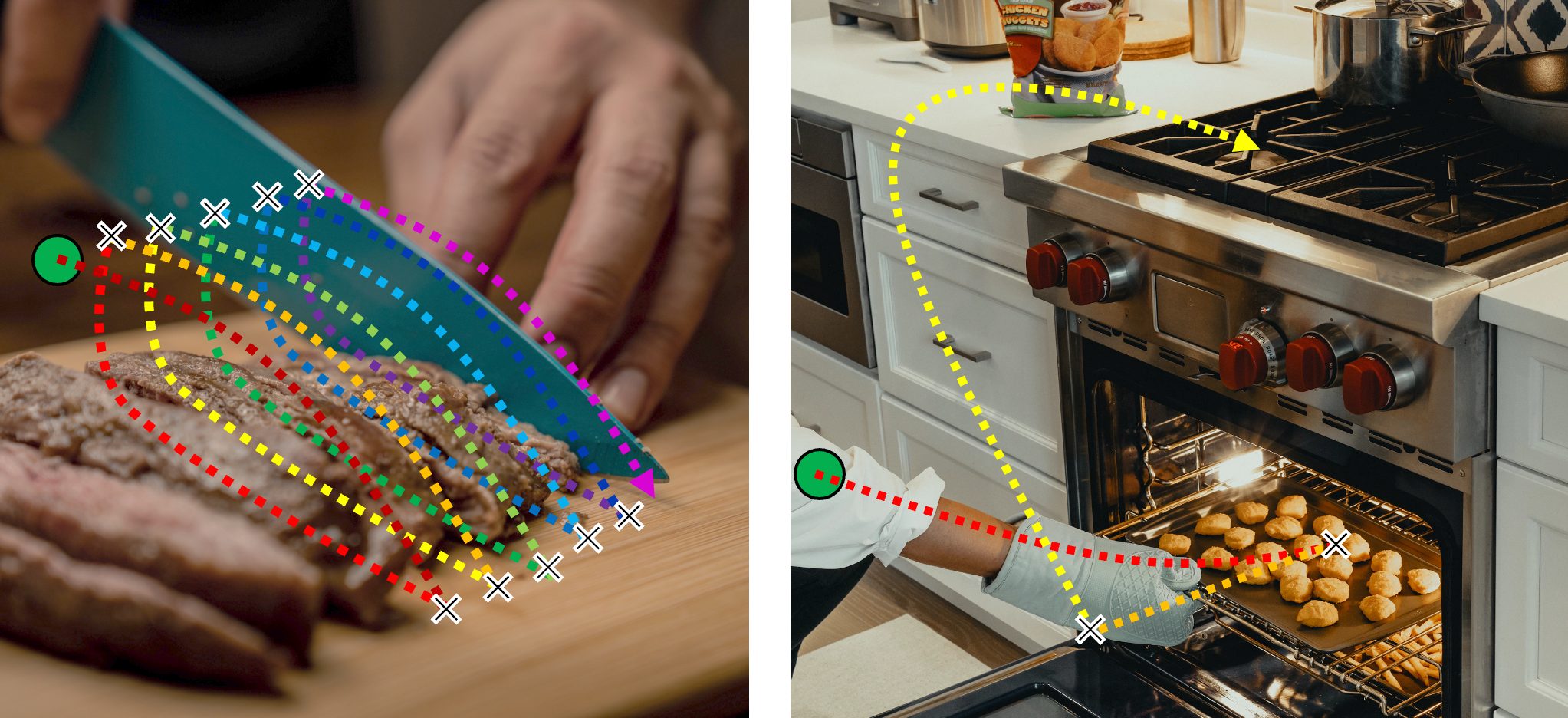}
\caption{
Discontinuous motions found in everyday life. Initial position of the motion is shown by the green dot while the transition between continuous motions is shown by $\boldsymbol \times$.
}
\label{fig:introduction-img}
\end{figure}

With the increasing usage of robotics in various domains, robots must be able to perform tasks with varying complexity and precision such as shown in \figref{fig:introduction-img}. 
These precise tasks often utilizes discontinuous motions where sudden change of direction or velocity happens within the motion \cite{ravankar2018path, qureshi2019motion} connected by a short-pause which is necessary to prevent overshoot especially in fragile environments. 
In addition to utilizing discontinuous motion, robots also need the intelligence to consider the environment's state received through sensor inputs to properly perform its task which in recent years comes in the form of imitation learning (IL) \cite{choudhury2018data}. 

A well known method to generate those robotic motor behavior is the motion generation method called Dynamic Movement Primitives (DMP) \cite{ijspeert2013dynamical}. Beside being able to represent a motor movement with a set of parameters, motion generation through DMP also boasts its stability as it is based on the well-established spring-damper model. Through the use of separate scalable and temporal parameters, DMP also features scalability in both spatial and temporal domain which adds more flexibility to the motion generation process. Recent trend which improves upon DMP is the prediction of DMP components such as its parameters or forcing function through deep learning based methods \cite{pahic2020training} which allows the generation of motion represented by DMP with regards to input data. 

Nevertheless, an issue exist that would prevent accurate generation of discontinuous motion using a DMP based imitation learning which stems from how DMP represents a motion.
To accurately represent a motion using DMP, multiple weighted Gaussian basis functions are combined to construct a forcing function that alters the motion's acceleration. However, discontinuous motions with multiple accelerations and decelerations require more basis functions and larger basis function weights, making them unsuitable for DMP-based imitation learning models as it would expand the feature size by a large degree. 
For this problem, we found that by decomposing the discontinuous motion, simpler primitives easier to represent with DMP are produced, which ultimately resulted in a shorter or at least equal length of feature size, with an additional benefit of a more accurate discontinuous motion generation.

Based on the idea mentioned above, we propose a novel deep auto-encoder based neural network which produces variable-length segmented motion by implementing:
\begin{enumerate*}
    \item \textbf{multiple DMP parameters predicting network architecture} to support prediction of varying number of segments,
    \item \textbf{double-stage decoder network} to capture similarities that can be found in-between segments of the same data, and
    \item \textbf{number of segments predictor} to limit the final number of segments.
\end{enumerate*}
The proposed method is evaluated using two artificially generated data (object cutting and pick-and-place) and a real data (object cutting). 
Based on the evaluation result, the proposed method has shown to outperform comparison method in generalization capability, task achievement, and data-efficiency.

The main contributions of this paper are:
\begin{enumerate*}
    \item proposing a data-efficient deep learning based motion generation method which is effective against long-horizon robotic tasks requiring discontinuous motion,
    \item implementing motion sequencing within an imitation learning framework, and
    \item an empirical study of the proposed method and existing DMP-based deep imitation learning framework on the task of object cutting and pick-and-place with varying difficulties and dataset amount.
\end{enumerate*}

\section{RELATED WORK}

\subsection{Path Continuity}
In order to identify the problem with representing discontinuous motion in robotics, it is best if we know exactly the characteristics of a discontinuous motion which is closely related to the topic of path continuity \cite{ravankar2018path}. 
Path continuity can be described by 2 types:
\begin{enumerate*}
    % \item 
    \item Geometric continuity which ensures the endpoints between segments meets with equal tangent vector's direction, and
    \item Parametric continuity which ensures the endpoints between segments meets with equal tangent vector's direction and magnitudes.
\end{enumerate*}
Based on that definition, discontinuous motion in robotics can be defined as a combination of segments of motion which endpoints
\begin{enumerate*}
    \item have different direction including curves, and
    \item have different velocity,
\end{enumerate*}
with the requirement of having the endpoints meet being a necessity, as an executed motion couldn't just suddenly displace to another position without any translation.

\subsection{Motion Representation through DMP Sequencing}
An ongoing trend within the DMP domain is the process of sequencing multiple movement primitives in order to learn motor skills which requires complex motion \cite{manschitz2014learning, cho2020learning, li2017reinforcement}. 
Through primitives sequencing, complex motions could be reconstructed more accurately as compared to motion reconstruction by a single DMP. Past works on this topic however are mostly implemented within a reinforcement learning framework, which is often dangerous to be used in the real world.

Advancements in this research area comes in the form of methods to seamlessly join the primitives thus reducing sudden changes in velocity \cite{kulvicius2011joining, nemec2012action}. This improvement however, would only be strictly required in connecting primitives where reconstruction of velocity profile is strictly required, which is not the case if the segments are connected at zero velocity as shown in this work.

\subsection{Deep Learning Implementation of DMP}
Recent developments in DMP research expands on DMP application in the deep learning domain \cite {schaal2003control} which can be categorized into imitation learning (IL)\cite {osa2018algorithmic} and reinforcement learning (RL) \cite{sutton2018reinforcement}. 
In the IL domain, deep learning capability to process different kind of input types can be utilized to associate various input state into components of DMP used to reconstruct the motion, resulting in a state-motion mapping framework. 
Some works which use this approach include Deep-DMP by Pervez et al. which maps input image into DMP's forcing terms \cite{pervez2017learning}, and Image-to-Motion Encoder Decoder Network by Pahic et al. which utilizes latent space to map input image into its corresponding DMP parameters \cite{pahic2018deep, pahic2020training}. 
While this approach does improve upon DMP's inability to process other type of inputs, naive prediction of DMP's components often resulted in inaccurate motion reconstructions which appears especially with complex motions that requires a large size of the DMP components to represent.

DMP's implementation in the RL domain on the other hand, while also predicts DMP's components, does so multiple times as the action produced by the control policy \cite{li2017reinforcement, stulp2011hierarchical, stulp2012reinforcement}. This behavior of generating multiple primitives and executing them sequentially in essence is the behavior of primitives sequencing which is an effective approach on generating a complex DMP-based motion \cite{meier2012movement, lioutikov2016learning}. Stulp et al. investigated on this issue on their work which utilizes DMP-based RL for robust manipulation \cite{stulp2011hierarchical, stulp2012reinforcement}. This application of primitives sequencing however hasn't been much investigated in the IL domain which could utilize demonstration data to replace RL's need for explorations.

\section{PRELIMINARY}
\label{prelim}

\subsection{Dynamic Movement Primitives}
\label{dmp}
DMP is based on the spring damper model \cite{ijspeert2013dynamical} shown by the following differential equations,
\begin{gather}
    \tau\dot{\boldsymbol z} = \boldsymbol \alpha_z(\boldsymbol \beta_z(\boldsymbol g-\boldsymbol y)-\boldsymbol z)+\boldsymbol F(x), \label{eq:dmp_diff_1} \\
    \tau\dot{\boldsymbol y} = \boldsymbol{z}, \label{eq:dmp_diff_2}
\end{gather}
where $\tau$ is a time constant responsible for temporal scaling, $\boldsymbol \alpha_{z} \in \mathbb{R}^d$ and $\boldsymbol \beta_{z} \in \mathbb{R}^d$ are positive constants which when assigned the appropriate value ($\boldsymbol \beta_{z} = \boldsymbol \alpha_{z} /4$), causes the dynamical system to be critically damped, $\boldsymbol y \in \mathbb{R}^d$ and $\boldsymbol g \in \mathbb{R}^d$ are the current position and the goal position which is responsible for the spatial scaling, $\boldsymbol F \in \mathbb{R}^d$ is a forcing function of phase $x \in \mathbb{R}$ which is responsible for altering the dynamical system trajectory, phase $x$ is a component used to avoid explicit time dependency within the dynamical system, and finally $d$ is degree-of-freedom of the dynamical system.
In this research we used the discrete implementation of DMP which  attractor system is defined as,
\begin{gather}
    \boldsymbol F(x) = \frac{\sum_{i=1}^{N}\psi_i(x)\boldsymbol w_i}{\sum_{i=1}^{N}\psi_i(x)}x(\boldsymbol g-\boldsymbol y_0), \label{eq:forcing_function} \\
    \psi_i(x) = \exp{(-\frac{1}{2\sigma_i^2}(x-c_i)^2)}, \label{eq:psi}
\end{gather}
where $\psi_i(x)$ are $N$ exponential basis functions with $\sigma_i \in \mathbb{R}$ and $c_i \in \mathbb{R}$ acting as the width and centers of the basis functions weighted by $\boldsymbol w_i \in \mathbb{R}^d$. Forcing function \equref{eq:forcing_function} is modulated by $(\boldsymbol g-\boldsymbol y_0)$ which corresponds to the scaling properties of the model by the initial position $\boldsymbol y_0$ and goal position $\boldsymbol g$, and the phase $x$ which vanishes the forcing term when the goal $\boldsymbol g$ is reached. 

Given this forcing function equation, we can see that the number of basis function and its corresponding weight is the one mostly responsible for defining the resulting force profile which depends on the complexity of the motion, such as
\begin{enumerate*}
\item \textbf{how often accelerations and decelerations happen}, and
\item \textbf{the intensities of the acceleration and decelerations}.
\end{enumerate*}
Based on this complexity criteria, a complex motion would require more basis functions with large weight value to accurately reconstruct using DMP. 
\begin{figure}[t]

    \begin{minipage}[b]{1.\linewidth}
    \centering
    \includegraphics[width=1.0\hsize]{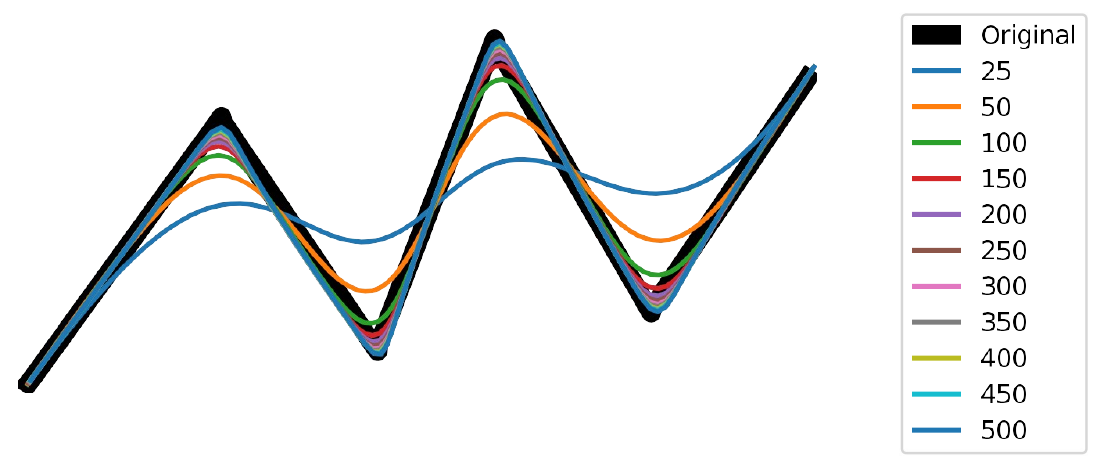}
    \subcaption{}
    \end{minipage}
    \begin{minipage}[b]{0.49\linewidth}
    \centering
    \includegraphics[width=1.0\hsize]{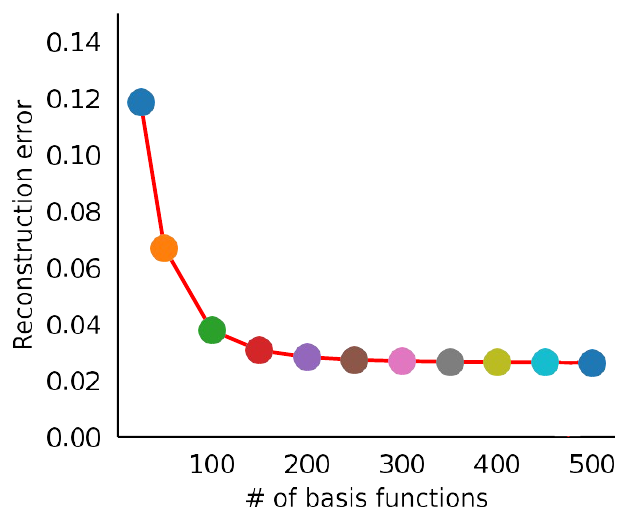}
    \subcaption{}
    \end{minipage}
    \begin{minipage}[b]{0.49\linewidth}
    \centering
    \includegraphics[width=1.0\hsize]{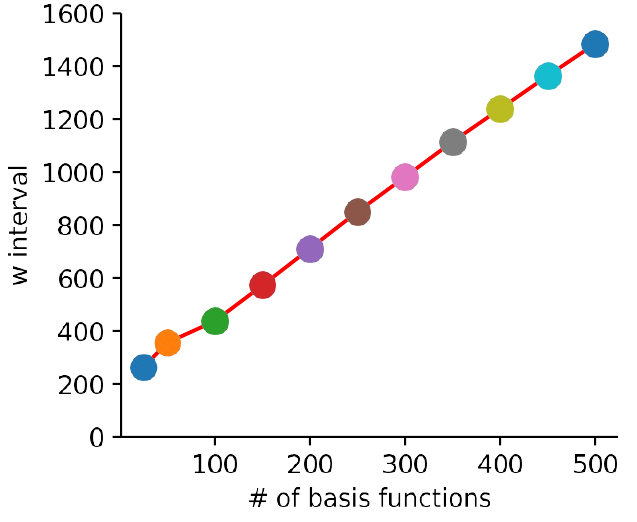}
    \subcaption{}
    \end{minipage}
    \caption{
        (a) DMP-reconstructed discontinuous motion with varying number of basis functions, (b) Reconstruction error decreases with the increase of basis functions but some error is still available even with a large number of basis functions, (c) DMP basis functions weight interval increases with more basis functions used.
    }
    \label{fig:jagged_reconstruct}
\end{figure}

\subsection{Convolutional Image-to-Motion Encoder Decoder\\Network (CIMEDNet)}
\begin{figure}[t]
\centering
    \includegraphics[width=1.0\hsize]{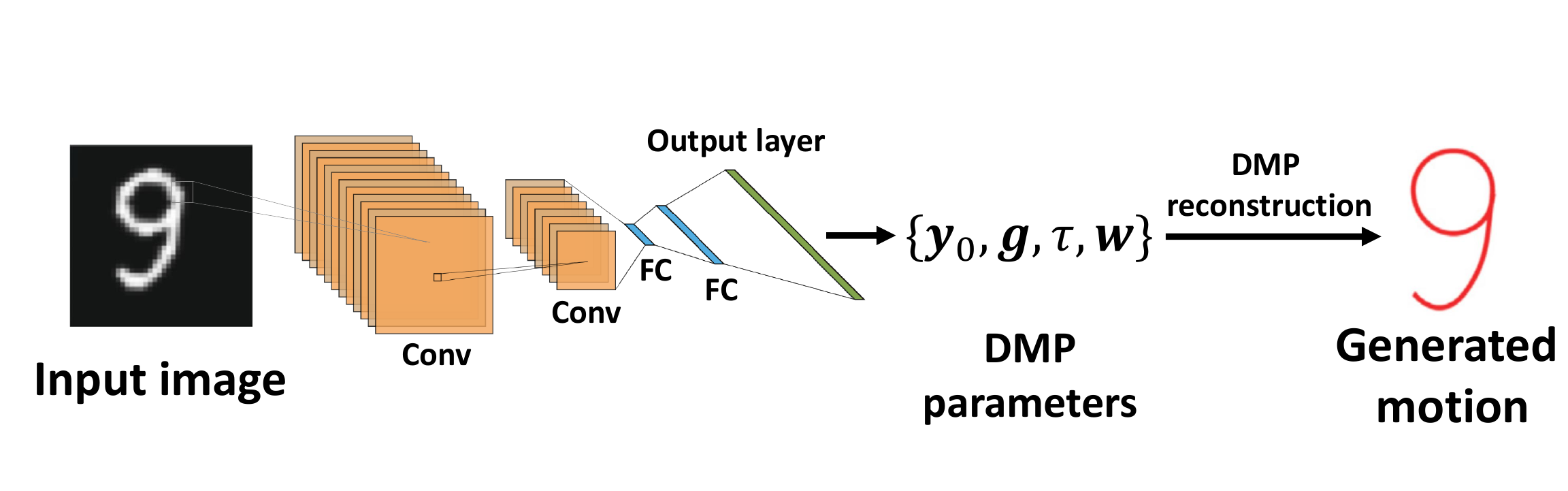}
\vspace{-7mm}
\caption{
CIMEDNet generating a DMP-based motion for a tracing task.
}
\label{fig:cimednet}
\end{figure}

Previous work done by Pahic et al. called Convolutional Image-to-Motion Encoder Decoder Network (CIMEDNet) \cite{pahic2020training} utilizes latent spaces within a deep encoder decoder network to map an image input to its corresponding DMP parameters encoded motion.
Four DMP parameters ($\{\boldsymbol y_0, \boldsymbol g, \tau, \boldsymbol w\}$) are predicted by CIMEDNet used to define a DMP attractor landscape, as shown in \figref{fig:cimednet}.

In their implementation, Pahic et al. normalized their predicted DMP parameters to avoid large value difference between parameters. However, normalizing DMP parameters with a large interval can cause sensitivity issue on its denormalization, requiring the deep learning model to be highly accurate in its predictions. Additionally, when a motion requires a lot of basis functions to reconstruct, the neural network architecture would require more neurons in the output layer to predict the additional basis function weights. 
Both of these issue are visualized in \figref{fig:jagged_reconstruct}.

Therefore, to summarize the issues that appear when we try to predict the DMP parameters for a discontinuous motion using CIMEDNet,
\begin{enumerate*}
\item the size of the network output layer will be large because of the large number of basis functions required, and
\item the need for a very accurate prediction because of the large value interval of the DMP parameters.
\end{enumerate*}
Both of these issues resulted for the need of a very accurate model representation, which for a deep learning based model means that a large amount of good quality training data is required.

\begin{figure*}[t]
\centering
    \includegraphics[width=1.0\hsize]{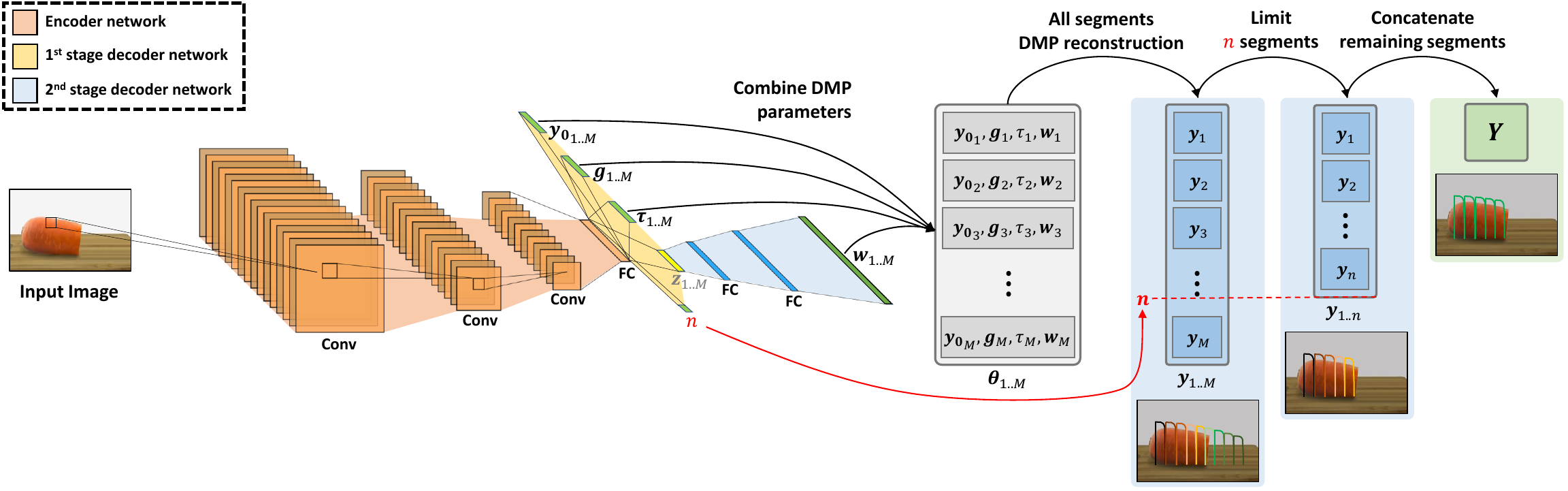}
\caption{
DSDNet motion reconstruction process. (i) Multiple motion are reconstructed by DMP from sets of DMP parameters predicted through (ii) double-stage decoder network architecture for data-efficiency. In order to produce the variable-length segments motion, (iii) number of segments predictor predicts $n$ which is utilized to limit the used segments in the final motion.
}
\label{fig:dsdnet}
\end{figure*}

\section{PROPOSED METHOD}

This section explains the idea and details of the proposed method visualized in \figref{fig:dsdnet}. As explained in the previous section, the problem of predicting a discontinuous motion's DMP parameters goes back to the complex force profile of a discontinuous motion given the complexity criteria described in \sref{dmp}. Thus, in order to solve upon the issues that appears when reconstructing a discontinuous motion with DMP, the most obvious solution would be to learn multiple simpler motion with low complexity instead of learning a complex motion. This solution provides several main issues within its implementation such as
\begin{enumerate*}
    \item inconsistencies of segments length,
    \item requiring a large amount of data to learn all combinations of DMP parameters between all segments, and 
    \item how to decide the final length of segments
\end{enumerate*}
To address this issue, we proposed a variable-length motion generation method by implementing 
\begin{enumerate*}
    \item \textbf{multiple DMP parameters predicting network architecture},
    \item \textbf{double stage decoder network}, and
    \item \textbf{number of segments predictor}.
\end{enumerate*}

\subsection{Deep Segmented DMP Network (DSDNet)}
\label{DSDNet}

\subsubsection{Multiple DMP parameters predicting network architecture}
DSDNet uses latent space within an encoder decoder network to predict a fixed $M \in \mathbb{N}$ sets of DMP parameters. 
The first decoder network in DSDNet predicts $\{n, \{\boldsymbol y_{0_m}\}^{M}_{m=1}, \{\boldsymbol g_{m}\}^{M}_{m=1}, \{\tau_{m}\}^{M}_{m=1}, \{\boldsymbol z_{m}\}^{M}_{m=1}\}$ where $n \in \mathbb{N}$ is the predicted number of segments, and $\boldsymbol z \in \mathbb{R}^M$ is the latent space representation of $\boldsymbol w$ which will goes through the next decoder network.

\subsubsection{Double-stage decoder network}
Before $\boldsymbol z$ goes through the second decoder network, we vectorize $\boldsymbol z$ with the number of batch $b \in \mathbb{N}$ as $\boldsymbol z \in \mathbb{R}^{b,M} \rightarrow \boldsymbol z \in \mathbb{R}^{b \times M}$, thereby training the second decoder network for each latent space value separately.

Through this implementation of double-stage decoder network, not only similarity in basis function weights that can be found between different data is learned, similarity between different segments of the same data could also be learned, thus allowing the second decoder network to learn the basis function weights combinations with more data compared to other DMP parameters. 
Even though this implementation only covers basis function weights, it is still an essential part to our proposed method as basis function weights normally takes the largest ratio within a set of DMP parameters.

\subsubsection{Number of segments predictor}
Sets of DMP parameters $\Theta$ can be generated through the combination of values from $\boldsymbol y_0$, $\boldsymbol g$, $\tau$, and the resulting output $\boldsymbol w \in \mathbb{R}^{b \times M \times N} \rightarrow \boldsymbol w \in \mathbb{R}^{b,M,N}$ of the second decoder network vectorized back into its batch form. Through modification of Eq. \eqref{eq:dmp_diff_1} to support batch calculation by replacing scalar into vector calculation, parallel computation of DMP reconstruction is enabled by treating the segments as a batch producing $M$ reconstructed motion $\boldsymbol y \in \mathbb{R}^{M, t, d}$. Finally, a subset of the reconstructed motion limited by the predicted number of segments $n$ is concatenated to produce the final motion $\boldsymbol Y \in \mathbb{R}^{n \times t, d}$. 

A fixed size of dataset with $M$ segments is required for this implementation, which is not the case for discontinuous motions with inconsistent number of segments, an issue which we addressed through segments padding. 
To reduce outlier values for the padding, we pad the missing $i^{th} \in \mathbb{N}$ DMP parameters segment with the average parameter value across all data where $i^{th}$ segment is available. 
The original number of segments $n \in \mathbb{N}$ is also taken as label data.
Finally, by defining the maximum number of segments $M$, padding the data with segments less than $M$, and also defining the number of segments $n$ for each data, a dataset with a fixed size is produced.

\subsection{Loss Function}
In our proposed method, the length of the final motion is not only altered by each $\tau$ value for each segments, but is also limited by the number of segments $n$. Given these limitations, we chose the comparison between DMP parameters of all segments while also comparing the number of segments $n$ as our loss function to optimize our encoder-decoder network defined as follows,
\begin{align}
    E = \sum\nolimits_{i=1}^{M} \Big(& \|{\boldsymbol y_0}_{i}-\hat{{\boldsymbol y_0}}_{i}\|^2 + \|\boldsymbol g_{i}-\hat{\boldsymbol g}_{i}\|^2 + (\tau_i - \hat{\tau}_i)^2 + \nonumber \\ 
    &\sum\nolimits_{j=1}^{N} \|\boldsymbol w_{i,j}-\hat{\boldsymbol w}_{i,j}\|^2 \Big) + \|{\boldsymbol n}-\hat{{\boldsymbol n}}\|^2,
\label{eq:cost_function}
\end{align}
where $\hat{\boldsymbol y_0}$, $\hat{\boldsymbol g}$, $\hat{\tau}$, $\hat{\boldsymbol w}$, and $\hat{n}$ corresponds to predicted initial position, predicted goal position, predicted $\tau$ value, predicted basis functions weight, and predicted number of segments respectively.

\section{EVALUATION: ARTIFICIAL DATA}
\label{eval-artificial}
Experiments utilizing artificial data are conducted to evaluate the proposed method on several different performance measures, including:
\begin{enumerate*}
    \item \textbf{Generalization capability} is measured between the generated motion and the task-solution of the test data by Root Mean Squared Error (RMSE) with heuristical sampling to handle difference in vector length and also Dynamic Time Warping (DTW) which is a distance measure of spatial information between two vectors regardless of temporal differences \cite{sakoe1978dynamic}. 
    \item \textbf{Task achievement} evaluation depends on each task definition of success. 
    \item Lastly, \textbf{data-efficiency} is evaluated through both generalization capability and task achievement when the model is trained using limited number of data. 
\end{enumerate*}
CIMEDNet is used as a comparison DMP-based motion generation method.

To ensure the fairness of both models, the amount of basis functions in the DMP parameters used to train CIMEDNet(=) is the same as DSDNet's maximum number of segments $M$ multiplied by the number of basis functions used for the DMP parameters of each segment. In addition, another CIMEDNet model called CIMEDNet(+) trained using DMP parameters with more basis functions to reconstruct the motion accurately with a trade-off of having a larger output layer size is prepared.

For the artificially generated data, 2 types of data are generated
\begin{enumerate*}
    \item periodic discontinuous motion data from the object cutting task, and 
    \item non-periodic discontinuous motion data from the randomized pick-and-place task.
\end{enumerate*}
Note that even with the task utilizing periodic discontinuous motion, some adjustment still need to be made on the positions parameters of each segment made possible by utilizing a deep learning model.
The task evaluation process is also designed to consider the adjustable position parameter according to the input image, thus rendering the usage of conventional periodic DMP to be challenging.

\subsection{Object cutting}

\subsubsection{Task setting}

Given a snapshot of the initial state of an object with differing shape and size, the motion generation method is tasked with generating a periodic cutting motion appropriate for the current object. 
For this evaluation, degree-of-freedom (DoF) of the generated motion is limited to 2 axis which are the up-down movement and right-left movement. The distance between cuts and the initial position are both fixed to simplify the task. 

To evaluate the task achievement, we counted the total number of successful cuts for each model on the test data. A cut is considered successful if it lifts the knife above the object and cuts through a threshold line on the bottom of the object. To evaluate the data-efficiency performance of the proposed method, another set of models trained on a smaller number of dataset is prepared. 

\subsubsection{Dataset creation}

We started with the generation of randomized-shape polygon which is used for both input image and cutting motion generation.

For the input image, the polygon is projected into a viewport as white color on top of the black background, simulating an image acquired from a depth camera, or an image passed through a image segmentation process.

A heuristic algorithm is used to place a pre-defined cutting motion which include a downward motion to the object, and an up-right curving motion which goes to the next cutting position. Utilizing the polygon's coordinates, the pre-defined cutting motion is scaled vertically to reach a certain constant distance above the highest horizontally-closest coordinate.
To produce the next cutting segment, we repeat the process above while continuing from the last segment's position. The whole cutting motion is finished with a downward motion as the up-right curving motion is unnecessary.

With the segments of cutting motion placed by the process above, we finally generate the DMP parameters for each segments through locally weighted regression\cite{ijspeert2013dynamical} for DSDNet, while 1 set of DMP parameters is generated for CIMEDNet.
The total number of both downward and up-right curving segments is also collected to evaluate the network output.

1000 data is generated for sufficient-data evaluation, while another dataset with 100 data is generated for the limited-data evaluation. Both are split 70\% for training data, 20\% for validation data, and 10\% for test data.

\begin{figure}[t]
\centering
    \includegraphics[width=1.0\hsize]{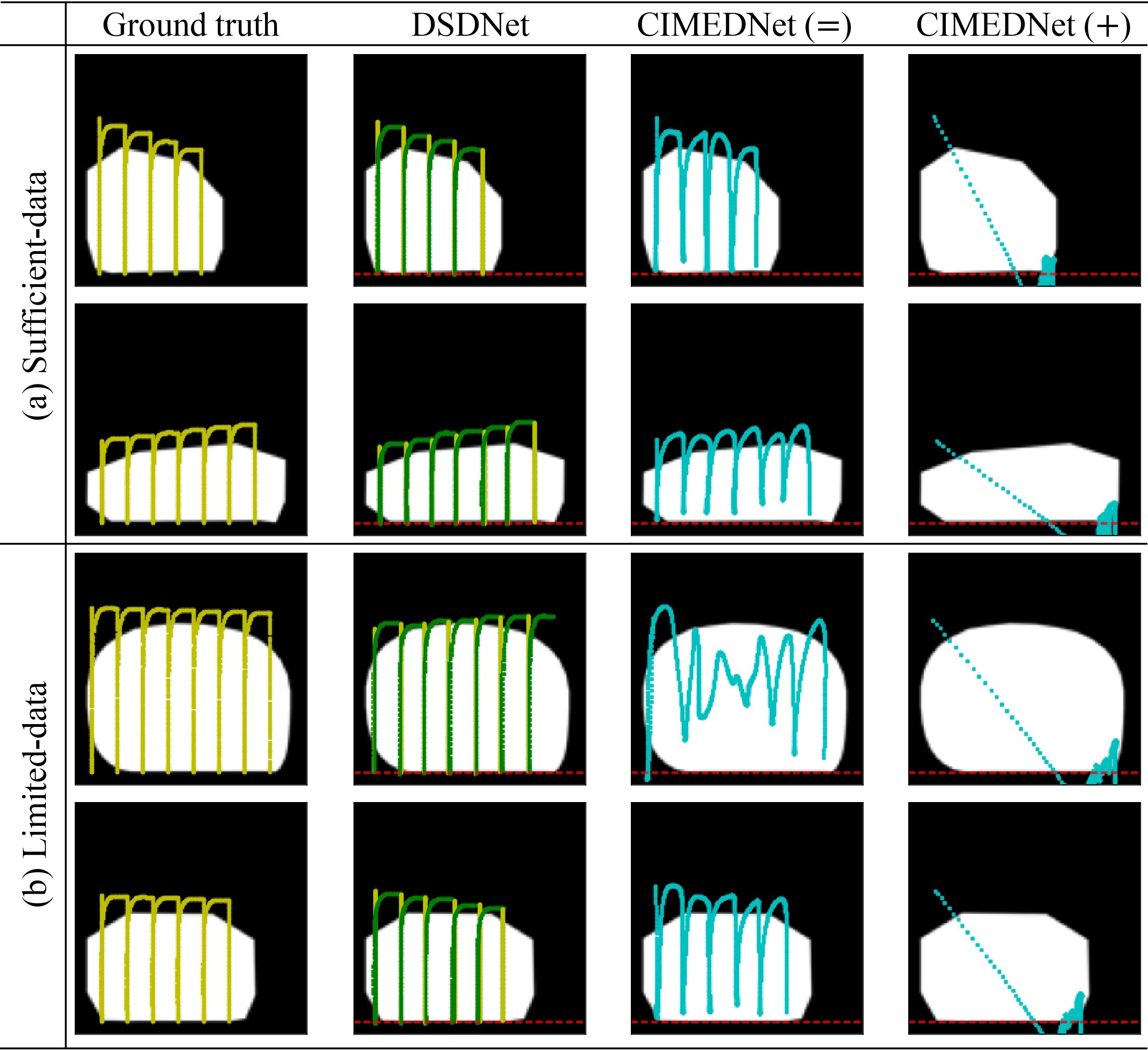}
\caption{
Projection of the cutting motion generated by all model trained on (a) sufficient-data, and (b) limited-data.
}
\label{fig:cut_projection_artificial}
\end{figure}

\begin{table}[t]
\caption{Object cutting task result evaluated through several evaluation metric. (a) corresponds to the results of models trained using sufficient-data, while (b) corresponds to the results of models trained using limited-data.}
\label{table:cutting}
\centering
\begin{tabular}{ c | c | c c c}
\hline
& Evaluation metric                    & DSDNet        & CIMEDNet(=)   & CIMEDNet(+)   \\
\hline
\multicolumn{1}{c|}{\multirow{3}{8pt}{(a)}} & RMSE                & \textbf{0.037}             & 0.145             & 0.628             \\
& DTW                 & \textbf{0.689}             & 1.178             & 11.934             \\
& Success-rate (\%)     & \textbf{98.598}             & 21.963             & 0             \\
\hline
\multicolumn{1}{c|}{\multirow{3}{8pt}{(b)}} & RMSE                & \textbf{0.111}             & 0.161             & 0.595             \\
& DTW                 & \textbf{1.900}             & 2.386             & 9.915             \\
& Success-rate (\%)    & \textbf{95.540}             & 14.085             & 0             \\
\hline
\end{tabular}
\end{table}

\subsubsection{Result}

\begin{figure}[t]
\centering
    \includegraphics[width=1.0\hsize]{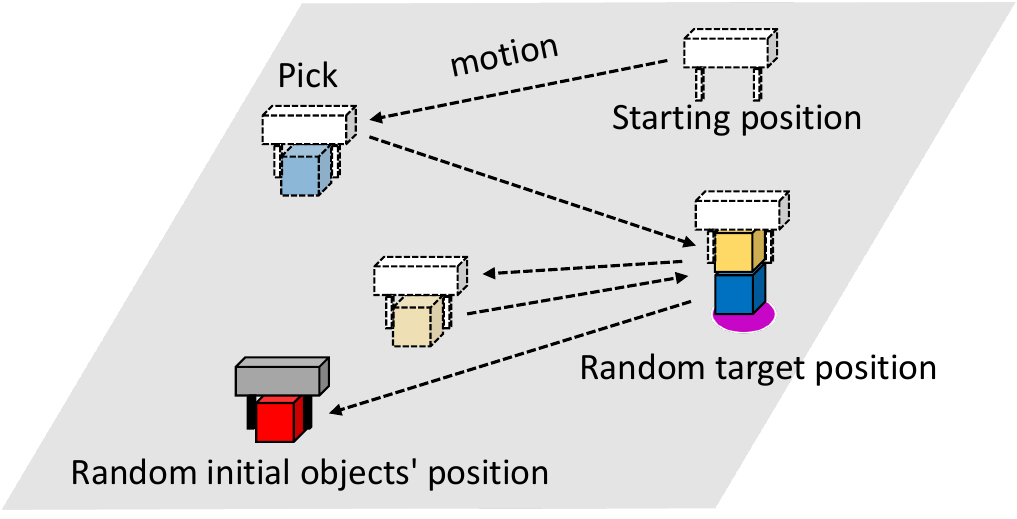}
\caption{
Examples of initial state for randomized pick-and-place task. Objects are shown by red, yellow, and blue squares and the target is shown by the purple circle.
}
\label{fig:pickplace_input}
\end{figure}

Projection of the generated motion to the input image trained on both sufficient-data and limited-data can be seen in \figref{fig:cut_projection_artificial}, while the task achievement result can be seen in \tableref{table:cutting}. From this result, we could see that DSDNet has shown the best performance in terms of generalization capability, task achievement, and data-efficiency compared to CIMEDNet-generated motion.

Motion generation by CIMEDNet seems to have difficulty generating an accurate discontinuous motion as it fully relies on the basis functions weight among all of the DMP parameters given the ratio of parameters in the network output layer. By failing to predict the basis functions weight accurately, especially for a discontinuous motion which requires high accuracy, the motion generated by CIMEDNet fails to reconstruct part of the motion which requires high intensity of acceleration or deceleration accurately as can be seen with the cutting motion making incomplete cuts. 

Failures in DSDNet on the other hand can mostly be attributed to its misprediction of the number of segments output rather than mispredicting the DMP parameters as can be seen in the third row of \figref{fig:cut_projection_artificial}. DSDNet doesn't have the same issues that troubles CIMEDNet as the basis functions weight predicted in DSDNet is less in quantity with smaller value interval. DSDNet also relies more on DMP's position parameters as each segment has an initial position and goal position parameters, reducing the responsibility of the basis functions weight on shaping the whole motion. This resulted in a final motion that could reach all short pauses within the discontinuous motion accurately.

\begin{table}[t]
\caption{Pick-and-place task result evaluated through several evaluation metric. (a) corresponds to the fixed version of the pick-and-place task, while (b) corresponds to the randomized version of the task.}
\label{table:pick_place}
\centering
\begin{tabular}{c | c | c c c}
\hline
& Evaluation metric                    & DSDNet        & CIMEDNet(=)   & CIMEDNet(+)   \\
\hline
\multicolumn{1}{c|}{\multirow{3}{8pt}{(a)}} & RMSE                & \textbf{0.404}             & 85797.703             & 247151.281             \\
& DTW                 & \textbf{31.917}             & 8422735.690             & 15707883.595             \\
& Success-rate (\%)     & \textbf{100}             & 0             & 0             \\
\hline
\multicolumn{1}{c|}{\multirow{3}{8pt}{(b)}} & RMSE                & \textbf{1.586}             & 25667.934             & 344754.573             \\
& DTW                 & \textbf{93.518}             & 2623380.032             & 29052772.126             \\
& Success-rate (\%)     & \textbf{77.778}             & 0             & 0             \\
\hline
\end{tabular}
\end{table}

\subsection{Pick-and-place}

\subsubsection{Task setting}
Given a snapshot of the initial state of an environment containing objects and a target position, the motion generation method is tasked with generating a 4 DoF motion for a gripper to pick the available objects, and place it on the target position. 3 DoF are responsible for the X-axis, Y-axis, and Z axis movement, while the last DoF is responsible for the gripper distance. In order to measure the generalization capability of this task, we evaluated the pick-and-place task on 2 different difficulties. For the fixed version of the task, the gripper are tasked on picking between 1 to 3 objects on a fixed position, and place them on a fixed target. On the randomized version, all positions of objects and target are randomized but does not overlap each other. Visualization of the task can be seen in \figref{fig:pickplace_input}. Task achievement for this task is calculated by the total number of successfully picked objects placed on or near the target position on the correct order of red $\rightarrow$ yellow $\rightarrow$ blue. 

\subsubsection{Dataset creation}
A simple pick-and-place environment is developed to generate data and later evaluate the task following the task setting above. 
150 data are generated for the simpler version of the task, while 9000 data are generated for the randomized version of the task. Both dataset are also split 70\% for training data, 20\% for validation data, and 10\% for test data.

\subsubsection{Result}
From the results of both version of the task shown in \tableref{table:pick_place}, DSDNet model can be seen to achieve far better results compared to both CIMEDNet models which completely fails on performing the task. 
CIMEDNet's difficulties in performing this task can be attributed to 2 reasons. 
For both version of the tasks, CIMEDNet is having difficulties on predicting the accurate DMP parameters given the discontinuous nature of the motion, even more for the randomized version of the task where all solutions will produce different DMP parameters especially for the basis functions weight.
DSDNet on the other hand, segments the discontinuous motion into simple primitives which can be represented using DMP with minimal basis functions, this resulted in DSDNet only requiring to predict the correct position parameters of each segments for the main issue of the task.

CIMEDNet's problem with this task also comes from the fact that DMP produces a curve-like motion unless enough basis functions are used, which resulted in failure by the gripper not completely grabbing the object. With the CIMEDNet(+) model which is supposed to be able to reconstruct the motion for the task properly, it faces the original issue that we are trying to tackle which is having a data with a large feature size, making it harder for the model to fit properly to the dataset.

\section{EVALUATION: REAL DATA}
\label{eval-real}

\begin{figure}[t]
\centering
    \includegraphics[width=1.0\hsize]{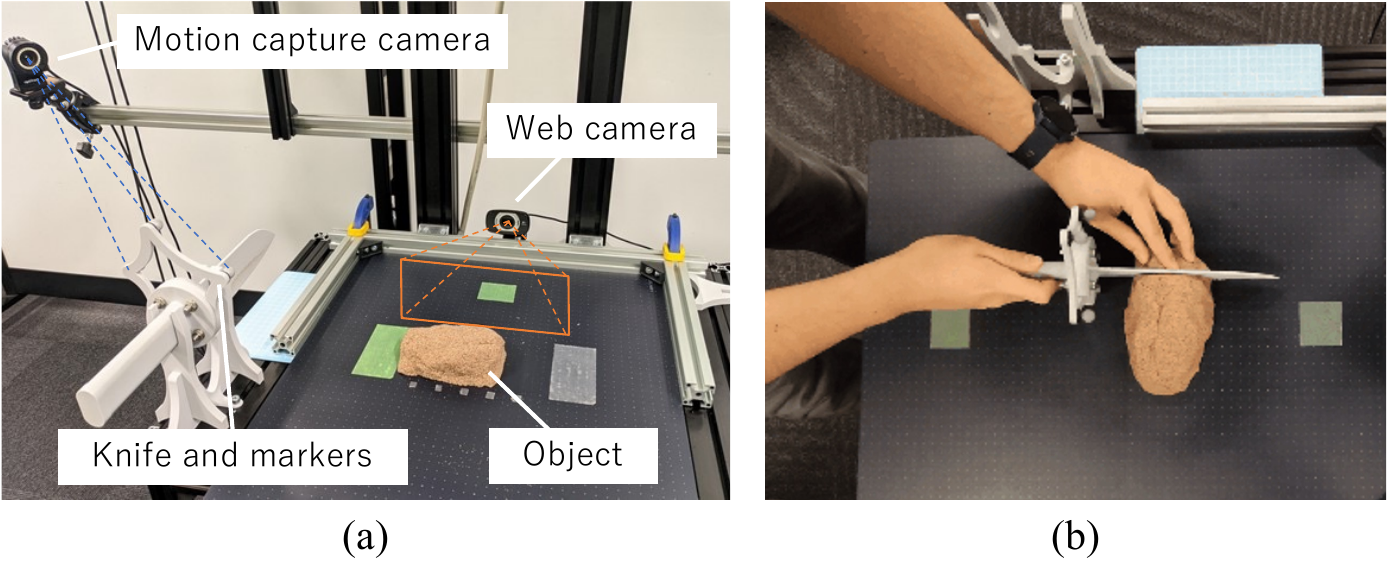}
\vspace{-7mm}
\caption{
The environment for collecting cutting data.
(a) The system captures the shape of the object using a web camera and the position of the knife through motion capture camera. A kinetic sand is used to simulate the cutting of an object.
(b) The top view of the environment. Human handles 3D-printed knife to collect real cutting data.
}
\label{fig:cut_real_env}
\end{figure}

\begin{figure}[tb]
\centering
    \includegraphics[width=1.0\hsize]{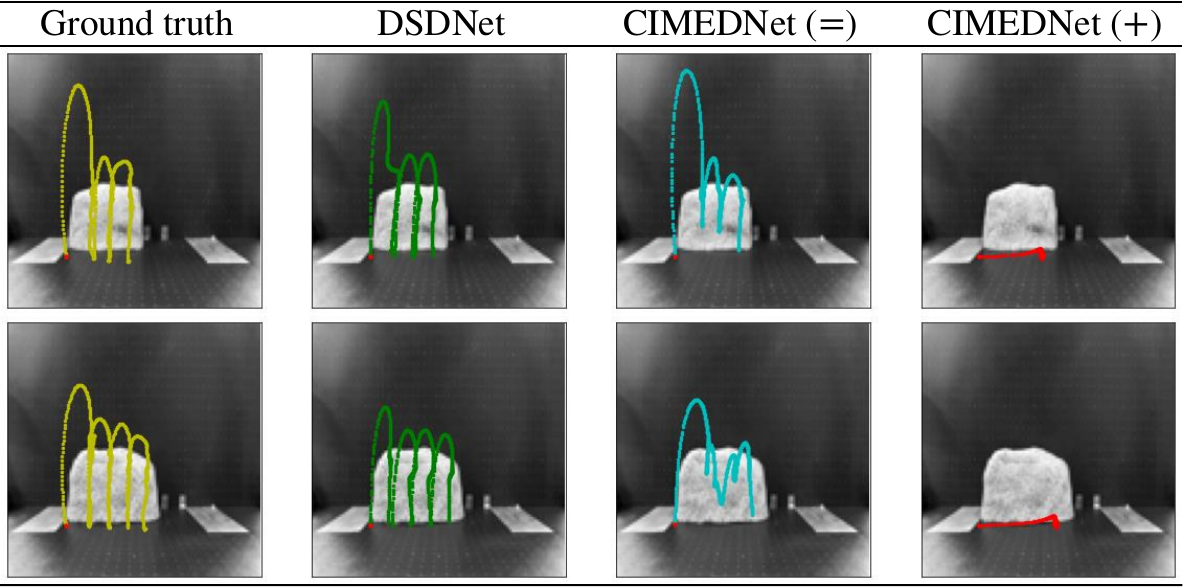}
\caption{
Projection of the cutting motion generated by all model trained on data collected through motion capture camera.
}
\label{fig:cut_projection_real}
\end{figure}

In this evaluation, the data-efficiency performance of both models and its ability to handle noisy data are evaluated in a simplified version of the object cutting task. Compared to the previous object cutting task, the cutting task for the real data evaluation is more simplified to reduce variations given the limited number of data. 
\subsubsection{Task setting}
Like the artificial object cutting task, the motion generation method is tasked with generating the appropriate cutting motion given an object image.
To simplify the task, the length of objects are limited to 3 types which would produce 3, 4, and 5 cuts. Though some additional difficulty is also added by using 3 DoF motion.
\subsubsection{Dataset creation}
Kinetic sand is used as a reusable cutting object to be cut by a 3D-printed knife model to support markers to be captured by a motion capture camera (OptiTrack: Flex13). Initial image state is captured by an off-the-shelf web camera.
The data collection environment can be seen in \figref{fig:cut_real_env}.

20 data are collected for each number of cuts, resulting in 60 total data.
This dataset is also split 70\% for training data, 20\% for validation data, and 10\% for test data.
\subsubsection{Result}
The projection of the generated motion of all models to the test image and the task achievement performance can be seen in \figref{fig:cut_projection_real} and \tableref{table:cutting_real} respectively. From the result shown in \figref{fig:cut_projection_real}, we can see that DSDNet could generate the appropriate cutting motion accurately given the limited number of data. The motion generated by CIMEDNet(=) on the other hand can be seen to almost resemble the original motion, but fails to capture the discontinuous motion properly thus failing to perform the task. Finally the motion generated by CIMEDNet(+) can be seen to be under-trained as the total number of data is very lacking to train all different combinations of basis functions weight.

\begin{table}[tb]
\caption{Object cutting task result of models trained using real data evaluated through several evaluation metric.}
\label{table:cutting_real}
\centering
\begin{tabular}{c | c c c}
\hline
Evaluation metric       & DSDNet        & CIMEDNet(=)   & CIMEDNet(+)   \\
\hline
RMSE (m)             & 0.016             & \textbf{0.014}             & 0.045             \\
DTW                 & \textbf{0.418}             & 0.762             & 1.827             \\
Success-rate (\%)     & \textbf{100}             & 13.636             & 0             \\
\hline
\end{tabular}
\end{table}

\section{DISCUSSION}
This section describes advantages, disadvantages, and the future direction of our proposed method.

Based on the evaluation results shown in the previous section, several confirmed advantages can be found within our proposed method as following:
\begin{enumerate*}
\item accurate discontinuous motion generation which has a high generalization capability to new states even with minimal amount of data,
\item a more task-friendly motion generation method done by segmenting the whole trajectory into multiple simple primitives which proves to be easier to reconstruct with DMP, and lastly
\item even though the proposed method is also based on deep learning, modifications proposed in this work resulted in a more data-efficient model which reduces usage burden.
\end{enumerate*}

We have several future directions to address the disadvantages found within the method,
\begin{enumerate*}
\item improving the generalization capability of the proposed method as currently it still has difficulty predicting the segments position with small amount of dataset. The inclusion of existing object detection methods may help with this issue.
\item improving the transition of position and velocity between DMP sequences, which would allow sequencing of continuous motion while also preserving its velocity profile, and
\item experimenting with other methods to represent DMP parameters, especially the basis function weights, which has proven to be difficult to learn when a deterministic method such as deep learning which requires large amount of data is used.
\end{enumerate*}

\section{CONCLUSION}
A new DMP-based motion generation method optimized for discontinuous motion is presented in this paper. The effectiveness of the proposed method is verified through long horizon motion generation tests utilizing both artificial and real data. The proposed method has shown to achieve better generalization capability and task achievement compared to previous methods while also reducing the amount of data required.

\bibliographystyle{IEEEtran}
\bibliography{reference}

\begin{thebibliography}{10}
\providecommand{\url}[1]{#1}
\csname url@samestyle\endcsname
\providecommand{\newblock}{\relax}
\providecommand{\bibinfo}[2]{#2}
\providecommand{\BIBentrySTDinterwordspacing}{\spaceskip=0pt\relax}
\providecommand{\BIBentryALTinterwordstretchfactor}{4}
\providecommand{\BIBentryALTinterwordspacing}{\spaceskip=\fontdimen2\font plus
\BIBentryALTinterwordstretchfactor\fontdimen3\font minus
  \fontdimen4\font\relax}
\providecommand{\BIBforeignlanguage}[2]{{%
\expandafter\ifx\csname l@#1\endcsname\relax
\typeout{** WARNING: IEEEtran.bst: No hyphenation pattern has been}%
\typeout{** loaded for the language `#1'. Using the pattern for}%
\typeout{** the default language instead.}%
\else
\language=\csname l@#1\endcsname
\fi
#2}}
\providecommand{\BIBdecl}{\relax}
\BIBdecl

\bibitem{ravankar2018path}
A.~Ravankar, A.~A. Ravankar, Y.~Kobayashi, Y.~Hoshino, and C.-C. Peng, ``Path
  smoothing techniques in robot navigation: State-of-the-art, current and
  future challenges,'' \emph{Sensors}, vol.~18, no.~9, p. 3170, 2018.

\bibitem{qureshi2019motion}
A.~H. Qureshi, A.~Simeonov, M.~J. Bency, and M.~C. Yip, ``Motion planning
  networks,'' in \emph{2019 International Conference on Robotics and Automation
  (ICRA)}.\hskip 1em plus 0.5em minus 0.4em\relax IEEE, 2019, pp. 2118--2124.

\bibitem{choudhury2018data}
S.~Choudhury, M.~Bhardwaj, S.~Arora, A.~Kapoor, G.~Ranade, S.~Scherer, and
  D.~Dey, ``Data-driven planning via imitation learning,'' \emph{The
  International Journal of Robotics Research}, vol.~37, no. 13-14, pp.
  1632--1672, 2018.

\bibitem{ijspeert2013dynamical}
A.~J. Ijspeert, J.~Nakanishi, H.~Hoffmann, P.~Pastor, and S.~Schaal,
  ``Dynamical movement primitives: learning attractor models for motor
  behaviors,'' \emph{Neural computation}, vol.~25, no.~2, pp. 328--373, 2013.

\bibitem{pahic2020training}
R.~Pahi^^c4^^8d, B.~Ridge, A.~Gams, J.~Morimoto, and A.~Ude, ``Training of deep
  neural networks for the generation of dynamic movement primitives,''
  \emph{Neural Networks}, vol. 127, pp. 121--131, 2020.

\bibitem{manschitz2014learning}
S.~Manschitz, J.~Kober, M.~Gienger, and J.~Peters, ``Learning to sequence
  movement primitives from demonstrations,'' in \emph{2014 IEEE/RSJ
  International Conference on Intelligent Robots and Systems}.\hskip 1em plus
  0.5em minus 0.4em\relax IEEE, 2014, pp. 4414--4421.

\bibitem{cho2020learning}
N.~J. Cho, S.~H. Lee, J.~B. Kim, and I.~H. Suh, ``Learning, improving, and
  generalizing motor skills for the peg-in-hole tasks based on imitation
  learning and self-learning,'' \emph{Applied Sciences}, vol.~10, no.~8, p.
  2719, 2020.

\bibitem{li2017reinforcement}
Z.~Li, T.~Zhao, F.~Chen, Y.~Hu, C.-Y. Su, and T.~Fukuda, ``Reinforcement
  learning of manipulation and grasping using dynamical movement primitives for
  a humanoidlike mobile manipulator,'' \emph{IEEE/ASME Transactions on
  Mechatronics}, vol.~23, no.~1, pp. 121--131, 2017.

\bibitem{kulvicius2011joining}
T.~Kulvicius, K.~Ning, M.~Tamosiunaite, and F.~Worg{\"o}tter, ``Joining
  movement sequences: Modified dynamic movement primitives for robotics
  applications exemplified on handwriting,'' \emph{IEEE Transactions on
  Robotics}, vol.~28, no.~1, pp. 145--157, 2011.

\bibitem{nemec2012action}
B.~Nemec and A.~Ude, ``Action sequencing using dynamic movement primitives,''
  \emph{Robotica}, vol.~30, no.~5, pp. 837--846, 2012.

\bibitem{schaal2003control}
S.~Schaal, J.~Peters, J.~Nakanishi, and A.~Ijspeert, ``Control, planning,
  learning, and imitation with dynamic movement primitives,'' in \emph{Workshop
  on Bilateral Paradigms on Humans and Humanoids: IEEE International Conference
  on Intelligent Robots and Systems (IROS 2003)}, 2003, pp. 1--21.

\bibitem{osa2018algorithmic}
T.~Osa, J.~Pajarinen, G.~Neumann, J.~A. Bagnell, P.~Abbeel, J.~Peters
  \emph{et~al.}, ``An algorithmic perspective on imitation learning,''
  \emph{Foundations and Trends{\textregistered} in Robotics}, vol.~7, no. 1-2,
  pp. 1--179, 2018.

\bibitem{sutton2018reinforcement}
R.~S. Sutton and A.~G. Barto, \emph{Reinforcement learning: An
  introduction}.\hskip 1em plus 0.5em minus 0.4em\relax MIT press, 2018.

\bibitem{pervez2017learning}
A.~Pervez, Y.~Mao, and D.~Lee, ``Learning deep movement primitives using
  convolutional neural networks,'' in \emph{2017 IEEE-RAS 17th international
  conference on humanoid robotics (Humanoids)}.\hskip 1em plus 0.5em minus
  0.4em\relax IEEE, 2017, pp. 191--197.

\bibitem{pahic2018deep}
R.~Pahi^^c4^^8d, A.~Gams, A.~Ude, and J.~Morimoto, ``Deep encoder-decoder
  networks for mapping raw images to dynamic movement primitives,'' in
  \emph{2018 IEEE International Conference on Robotics and Automation
  (ICRA)}.\hskip 1em plus 0.5em minus 0.4em\relax IEEE, 2018, pp. 5863--5868.

\bibitem{stulp2011hierarchical}
F.~Stulp and S.~Schaal, ``Hierarchical reinforcement learning with movement
  primitives,'' in \emph{2011 11th IEEE-RAS International Conference on
  Humanoid Robots}.\hskip 1em plus 0.5em minus 0.4em\relax IEEE, 2011, pp.
  231--238.

\bibitem{stulp2012reinforcement}
F.~Stulp, E.~A. Theodorou, and S.~Schaal, ``Reinforcement learning with
  sequences of motion primitives for robust manipulation,'' \emph{IEEE
  Transactions on robotics}, vol.~28, no.~6, pp. 1360--1370, 2012.

\bibitem{meier2012movement}
F.~Meier, E.~Theodorou, and S.~Schaal, ``Movement segmentation and recognition
  for imitation learning,'' in \emph{Artificial Intelligence and
  Statistics}.\hskip 1em plus 0.5em minus 0.4em\relax PMLR, 2012, pp. 761--769.

\bibitem{lioutikov2016learning}
R.~Lioutikov, O.~Kroemer, G.~Maeda, and J.~Peters, ``Learning manipulation by
  sequencing motor primitives with a two-armed robot,'' in \emph{Intelligent
  Autonomous Systems 13}.\hskip 1em plus 0.5em minus 0.4em\relax Springer,
  2016, pp. 1601--1611.

\bibitem{sakoe1978dynamic}
H.~Sakoe and S.~Chiba, ``Dynamic programming algorithm optimization for spoken
  word recognition,'' \emph{IEEE transactions on acoustics, speech, and signal
  processing}, vol.~26, no.~1, pp. 43--49, 1978.

\end{thebibliography}

\end{document}